\pdfoutput=1

\documentclass[11pt]{article}

\usepackage[]{acl}

\usepackage[colorinlistoftodos]{todonotes}

\usepackage{times,latexsym}
\usepackage{url}
\usepackage[T1]{fontenc}
\usepackage{caption}
\usepackage{subcaption}
\usepackage{float}
\usepackage{amsmath}
\usepackage[utf8]{inputenc} %
\usepackage[T1]{fontenc}    %
\usepackage{hyperref}       %
\usepackage{url}            %
\usepackage{booktabs}       %
\usepackage{amsfonts}       %
\usepackage{nicefrac}       %
\usepackage{lipsum}
\usepackage{amsmath}
\usepackage{fancyhdr}       %
\usepackage{graphicx}       %
\graphicspath{{media/}}     %
\usepackage{booktabs}
\usepackage{multirow}
\usepackage{bm}

\usepackage[english]{babel}
\usepackage[section]{placeins}

\usepackage{graphicx}

\usepackage{xspace,mfirstuc,tabulary}

\title{Unilogit: Robust Machine Unlearning for LLMs\\ Using Uniform-Target Self-Distillation}

\author{
  Stefan Vasilev\textsuperscript{\textnormal{1,2}}\thanks{Correspondence to: \href{mailto:stvasilev@ebay.com}{stvasilev@ebay.com}} \quad Christian Herold\textsuperscript{\textnormal{1}} \quad Baohao Liao\textsuperscript{\textnormal{1,2}} \\ 
  {\bf Seyyed Hadi Hashemi}\textsuperscript{\textnormal{1}} \quad
  {\bf Shahram Khadivi}\textsuperscript{\textnormal{1}} \quad
  {\bf Christof Monz}\textsuperscript{\textnormal{2}} \\
  \textsuperscript{\textnormal{1}}eBay Inc.\\
  \textsuperscript{\textnormal{2}}University of Amsterdam
}

\begin{document}
{
    \makeatletter\acl@finalcopytrue
    \maketitle
}

\begin{abstract}
This paper introduces Unilogit, a novel self-distillation method for machine unlearning in Large Language Models. Unilogit addresses the challenge of selectively forgetting specific information while maintaining overall model utility, a critical task in compliance with data privacy regulations like GDPR. Unlike prior methods that rely on static hyperparameters or starting model outputs, Unilogit dynamically adjusts target logits to achieve a uniform probability for the target token, leveraging the current model's outputs for more accurate self-distillation targets. This approach not only eliminates the need for additional hyperparameters but also enhances the model's ability to approximate the golden targets. Extensive experiments on public benchmarks and an in-house e-commerce dataset demonstrate Unilogit's superior performance in balancing forget and retain objectives, outperforming state-of-the-art methods such as NPO and UnDIAL. Our analysis further reveals Unilogit's robustness across various scenarios, highlighting its practical applicability and effectiveness in achieving efficacious machine unlearning.
\end{abstract}

\section{Introduction}
Large Language Models (LLMs) have advanced rapidly, becoming widely applicable in various settings \cite{DBLP:journals/corr/abs-2005-14165, DBLP:journals/corr/abs-2303-08774, DBLP:journals/corr/abs-2407-21783}. However, their increasing capabilities raise significant privacy risks, especially for individuals whose sensitive data may have been included in training. This information can become embedded within the model, making it susceptible to unintended exposure through memorization, adversarial exploits, membership inference (MIA), and model inversion attacks \cite{privacysurvey}.

To address these concerns, regulatory frameworks such as the \citet{GDPR2016} have been established to protect individual privacy and enforce the \textit{right to be forgotten}. Given that LLMs are subject to such regulations, the machine learning research community has increasingly focused on the emerging field of Machine Unlearning for LLMs \cite{wang2025rethinking, liu2024survey, jang-etal-2023-knowledge}, which aims to develop methods for selectively removing specific knowledge from models. This includes erasing sensitive information \cite{wang2025rethinking, patil2023can}, forgetting entire entities or facts \cite{ma-etal-2025-unveiling}, and removing harmful or biased information \cite{quark}.

In the machine unlearning framework, we define the full training dataset as a partition of two subsets: the \textbf{forget set}, which consists of the data to be unlearned, and the \textbf{retain set}, which contains the remaining knowledge that should be preserved after unlearning. An effective machine unlearning method aims to produce a model that successfully forgets the forget data, while maintaining the integrity of the retained knowledge. Specifically, the resulting unlearned model should satisfy the following key requirements: 1) Minimize the retention of information from the forget set; 2) Maintain high performance on the retain set; 3) Require less computational cost than retraining the model from scratch on the retain set; 4) Maintain inference efficiency, i.e., ensuring unchanged latency.

\begin{figure*}
    \centering
    \includegraphics[width=0.95\textwidth]{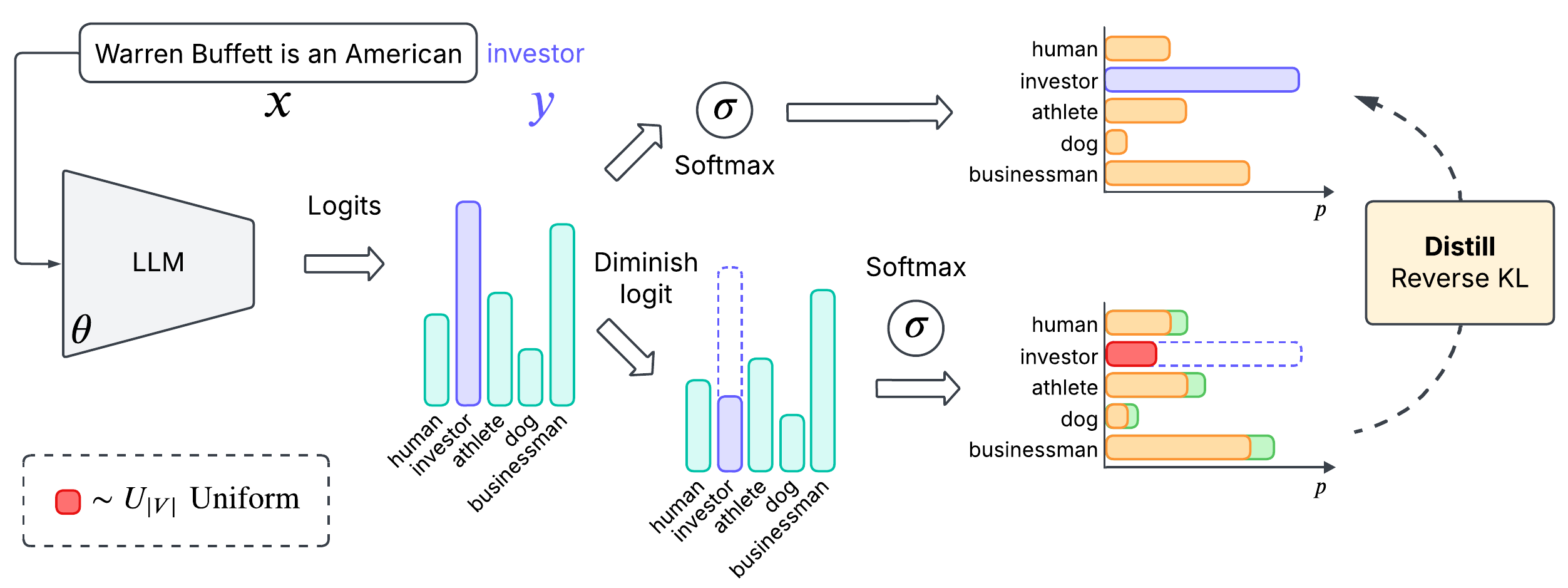}
    \caption{Overview of self-distillation unlearning in Unilogit: Starting with the output logits of the LLM, the target logit is diminished, so that after softmax the target token in the modified distribution has uniform probability. Soft labels are derived from the current model ($\theta$) outputs. Reverse KL divergence is the distillation objective.}
    \label{fig:enter-label}
\end{figure*}

Despite extensive research efforts \cite{wang2025rethinking, liu2024asurvey}, current unlearning methods still face significant challenges in achieving all these goals simultaneously. A major challenge in this domain is \textbf{catastrophic forgetting} \cite{zhang2024negative}, where the model suffers severe degradation in its ability to retain knowledge from the retain set while attempting to remove forget set knowledge. Additionally, unlearning methods must \textbf{balance forget and retain performance} \cite{wang2025flat}, as no existing technique can fully erase the forget set information, while preserving the original accuracy. To analyze this trade-off, researchers commonly visualize unlearning performance through \textit{Pareto frontiers} that plot forgetting effectiveness against retention performance across various hyperparameter sweeps \cite{zhang2024negative, dong2024undialselfdistillationadjustedlogits}. Finally, the balancing problem is related to the issues of \textbf{hyper-parameter tuning} and \textbf{robustness}, as each method and problem combination have their unique optimal set of hyperparameters \cite{yao-etal-2024-uniform, he2024towards}.

In this paper, we propose \textbf{Unilogit}, a self-distillation approach for unlearning. It generates targets from the output of the current model by assigning a uniform probability to the target token in forget samples and redistributing the remaining probability mass. Our method is inspired by \citet{dong2024undialselfdistillationadjustedlogits} and is driven by the question:
\begin{center} \label{rq:question}
\textit{Can we leverage existing information to refine the target distribution for forgetting?}
\end{center}
Unilogit offers a simple yet effective unlearning strategy that outperforms existing methods, demonstrating superior hyperparameter robustness and applicability across diverse scenarios, satisfying all key aforementioned unlearning requirements. Unlike prior techniques that introduce an extra hyperparameter in their loss \cite{zhang2024negative, dong2024undialselfdistillationadjustedlogits, wang2024rkldreversekldivergencebasedknowledge}, Unilogit achieves consistent forgetting performance without the extra tuning overhead. To validate our approach, we conduct in-depth auxiliary studies: (1) demonstrating that our soft labels and outputs are more accurate than those of other methods (Section \ref{sec:klevals}), and (2) ablation studies to assess the impact of key methodological components (Section \ref{sec:ablations}).

Our contributions are as follows:
\begin{itemize}
    \item We propose Unilogit, a novel method for machine unlearning that dynamically adjusts target logits to a uniform probability without additional hyperparameters, addressing catastrophic forgetting.
    \item We extensively evaluate Unilogit against state-of-the-art methods on various public benchmarks, demonstrating its robustness and effectiveness.
    \item We apply Unilogit to a real-life e-commerce use-case, showcasing its reliability in a practical scenario.
    \item We analyze Unilogit's self-distillation targets, demonstrating their accuracy compared to existing techniques through KL divergence studies, and perform ablation studies to assess the impact of key components, such as reverse KL divergence, highlighting the advantages of our approach.

\end{itemize}

\section{Background and Related Work}
\textbf{Background.}
Machine unlearning for LLMs focuses on removing specific knowledge from a trained model while preserving its overall performance. In this framework, the full training dataset is divided into two subsets: the forget set $D_f$, which contains the data to be unlearned, and the retain set $D_r$, which comprises the knowledge to be preserved. The primary objective is to approximate the performance of the golden \textit{retrained model} ($\theta_r$), which is trained solely on $D_r$. However, full retraining  on the retain set is often prohibitively expensive. Thus, machine unlearning seeks to provide a more efficient alternative. While \textit{exact unlearning} methods have been proposed \cite{ijcai2022p556, ding2024unified}, which fully retrain on $D_r$ on an algorithmic-level \cite{algorithmiclevel} to recover the exact behavior of the retrained model, these approaches require access to the complete retain set and are generally computationally expensive. In contrast, \textit{approximate methods} aim to closely approximate the retrained model's behavior through techniques such as fine-tuning \cite{yao2024large, zhang2024negative, pmlr-v132-neel21a}, prompting \cite{liu2024eco, pawelczyk2024incontextunlearninglanguagemodels}, or model editing \cite{veldanda2024editing, hase2023does}, offering a more scalable and efficient alternative.

\textbf{Parameter-tuning.} Among approximate methods, one prominent direction is \textit{parameter-tuning approaches}, which directly modify model parameters to achieve unlearning. We pursue this direction because these methods typically meet all unlearning requirements, preserving inference latency without demanding excessive training compute. Parameter-tuning methods frame unlearning as an optimization problem with two competing objectives: a \textit{forget objective} $\mathcal{L}_f$ that forces the model to unlearn specific knowledge and a \textit{retain objective} $\mathcal{L}_r$ that ensures performance on the remaining data is preserved. A generalized unlearning loss function typically follows this form:
\begin{align*}
    \operatorname*{argmin}_\theta \mathbb{E}_{(x_f, y_f) \in D_f}\left[\mathcal{L}_{f}(\theta, x_f, y_f, [\theta_o])\right] \, + \\
    \lambda \, \mathbb{E}_{(x_r, y_r) \in D_r} \left[\mathcal{L}_{r}(\theta, x_r, y_r, [\theta_o])\right] \,,
\end{align*}
where $D_f$ is the \textit{forget set}, $D_r$ is the \textit{retain set}, $\theta_o$ are the \textit{starting model weights} and $\lambda$ is a hyperparameter. Typically, approaches use $\theta_o$ in both objectives. The variation across methods lies primarily in how the forget loss $\mathcal{L}_f$ is designed. The retain objective $\mathcal{L}_r$ serves as a regularizer to mitigate catastrophic forgetting. Typically, either cross-entropy \cite{uniformce} or KL-divergence distillation from the starting model is used as retain loss with the latter usually performing better \cite{zhang2024negative, maini2024tofu}.

Notable parameter-tuning methods include \textbf{Gradient Ascent (GA)} \cite{jang-etal-2023-knowledge}, which suffers from instability, and \textbf{Negative Preference Optimization (NPO)} \cite{zhang2024negative}, which has emerged as a robust state-of-the-art method by introducing a controlled forgetting process. NPO uses a preference optimization-based loss function to mitigate the risk of catastrophic forgetting. Another relevant approach, \textbf{ME+GD} \cite{uniformce}, maximizes the entropy of the model’s predictions on the forget set by pushing the output probabilities towards a uniform distribution, preserving performance on the retain set using cross-entropy.

The concept of \textit{auxiliary models} in unlearning was introduced by \citet{eldan2023whosharrypotterapproximate} with their \textbf{"Who is Harry Potter" (WHP)} approach, which leverages a reinforced model fine-tuned on the forget set to inform the unlearning process. This idea has been further developed in \textit{distillation-based} unlearning methods, such as \textbf{RKLD} \cite{wang2024rkldreversekldivergencebasedknowledge} and \textbf{UnDIAL} \cite{dong2024undialselfdistillationadjustedlogits}. RKLD enhances the WHP approach by refining model reinforcement and distilling modified soft labels into the original model for targeted unlearning, while UnDIAL uses a self-distillation approach to adjust model logits for unlearning. Our approach is motivated by the ideas in UnDIAL, particularly in refining the process of generating effective soft labels for unlearning. 

Outside of the realm of text-based LLMs, the work most closely related to our approach is proposed by \citet{tang2024uniformsupervised}.
In their framework, they tackle unlearning by minimizing KL divergence between model outputs and a uniform distribution, then applying an MSE loss on the adjusted logits. While \citet{tang2024uniformsupervised} provide a general framework, particularly for weakly-supervised settings---where only limited or noisy supervision is available---and experiments with Computer Vision tasks, our method is specifically designed for LLMs. We achieve unlearning by optimizing reverse KL divergence at the categorical probability distribution level and dynamically updating target distributions based on the model’s latest outputs. These features enhance adaptability and unlearning effectiveness, setting our approach apart from existing methods.

For an extended discussion on related work, please refer to Appendix \ref{sec:app_related}.

\section{Methodology} \label{sec:methods}

In our self-distillation approach to unlearning, the central challenge is designing accurate soft targets that effectively guide the model toward forgetting. Ideally, the outputs of the retrained model $\theta_r$, would serve as the gold standard for distillation. However, since $\theta_r$ is unavailable in reality, we must approximate these targets through a principled and computationally efficient method that refines self-distillation for unlearning.

\textbf{Unilogit.} \label{sec:unilogit}
Inspired by recent advances in logit adjustment for unlearning \cite{dong2024undialselfdistillationadjustedlogits} and self-distillation, we propose \textbf{Unilogit}: a self-adjusting self-distillation method for machine unlearning.
It sets the target logit so that after the softmax operation, it is equal to a uniform distribution value, while preserving the logits for all other vocabulary entries. For a model output vocabulary $V$, output logit function $\bm{h}(x;\theta)$, parametrized by the \textit{current model}'s parameters $\theta$, and a one-hot label vector $\bm{t}$, we calculate the target logits $\bm{\tilde{h}}(x; \theta)$:  
\begin{align*}
    \bm{\tilde{h}}(x; \theta) = \, &
    (1-\bm{t})\, \bm{h}(x; \theta) \,+ \\ 
    & \bm{t} \, \log \sum\limits_{i \neq k}^{|V|}\frac{\exp({\bm{h}_i{(x;\theta})}) }{|V|-1}   
\end{align*}

Then, we calculate the soft label target distribution, where the target token label is going to be equal to a uniform probability ($=\frac{1}{|V|}$)  
\begin{equation*}
    \tilde{p}(y|x; \theta) = \mathrm{softmax}(\tilde{\bm{h}}(x; \theta))
\end{equation*}
For a detailed derivation, see Appendix \ref{appendix:derivation_unilogit}.

This design is grounded in the intuition that the current model $\theta$ and the retrained model $\theta_r$ should be relatively close in both parameter space and output distributions, given that the retain and forget sets originate from the same data distribution and the forget set is significantly smaller. Consequently, the non-target token logits of $\theta$ serve as a strong prior for approximating the output distribution of $\theta_r$. By explicitly setting the target token probability to a uniform value, we induce the desired unlearning effect while redistributing the lost probability mass according to this prior.

We adopt a uniform probability ($\frac{1}{|V|}$) for the target token as it represents a state of complete uncertainty, aligning with the goal of eliminating learned information about the forget token. This choice is also justified by prior work indicating that untrained models tend to produce nearly uniform output distributions \cite{tang2024uniformsupervised, uniformce}, making it a natural approximation of an untrained state. Importantly, this approach introduces no additional bias in determining the target token's probability, as we have no prior information about its true distribution post-unlearning.

Unilogit has two beneficial properties over UnDIAL: 1) it eliminates the need for a manually tuned hyperparameter $\gamma$ to scale down the target logit and 2) by explicitly setting the target probability to uniform, it dynamically adjusts the reduction factor in a self-consistent manner, ensuring stability and interpretability.

A crucial distinction between our approach and previous self-distillation-based unlearning methods \cite{dong2024undialselfdistillationadjustedlogits, wang2024rkldreversekldivergencebasedknowledge, tang2024uniformsupervised} is that we construct distillation targets from the current model parameters $\theta$ rather than the initial model parameters $\theta_o$. This choice is motivated by the assumption that a well-designed unlearning algorithm should progressively guide the model closer to the retrained model $\theta_r$. If this assumption holds, then at each unlearning step, the output distribution of $\theta$ should increasingly resemble that of $\theta_r$. By leveraging the latest model outputs as the basis for our self-distillation targets, we iteratively refine the approximation of $\theta_r$, leading to more accurate guidance for the unlearning process. This directly addresses the core question of our work (Question \ref{rq:question}), and we empirically validate this hypothesis in Section \ref{sec:klevals}.

Once we have our target distribution $\tilde{p}$, we can formulate the full training objective of Unilogit:
\begin{align*} 
    &\mathcal{L}_{\text{Unilogit+KL}}(\theta) = \\
    &\mathbb{E}_{(x_f, y_f) \in D_f} KL(p(y_f|x_f;\theta) \, || \, \tilde{p}(y_f|x_f;\theta)) \,+ \,\\ 
    \lambda \, 
    &\mathbb{E}_{(x_r, y_r) \in D_r} KL(p(y_r|x_r;\theta_o) \, || \, p(y_r|x_r;\theta))
\end{align*}
\label{eq:unilogit}

For the forget loss, we adopt reverse KL-divergence (RKL) due to its advantageous properties for unlearning. Unlike forward KL (FKL), which is mean-seeking, RKL is mode-seeking, thus penalizing cases where the model assigns high probability to an incorrect token for which the target distribution has assigned a low probability. This property is particularly well-suited for unlearning, as it strongly discourages the model from retaining high confidence in previously learned outputs. In Appendix \ref{sec:ablations}, we provide an ablation study demonstrating that RKL leads to superior unlearning performance compared to FKL.

Prior works \cite{wang2024rkldreversekldivergencebasedknowledge, wang2024beyondrkl} have also observed that RKL improves evaluation metrics at the cost of reduced generation diversity. However, in the unlearning setting, this tradeoff is acceptable, as we prioritize strong unlearning performance over minor diversity loss. Moreover, since our target distributions for non-target tokens remain largely consistent with the model’s original outputs, with the only major change occurring in the target token’s probability reduction, the overall impact on generation variance is expected to be minimal (see column Flu in Table \ref{tab:rwku}). Further supporting our approach, \citet{wu-etal-2025-rethinking} show that models optimized with RKL and FKL objectives eventually converge to similar distributions, reinforcing our decision to use RKL for this task.

\section{Experiments}

\textbf{Datasets and Models.}
We evaluate Unilogit on two public benchmarks. As a general unlearning scenario we choose \textbf{MUSE-News} benchmark (BBC News articles), for which we use Llama 2 7B \cite{touvron2023llama2openfoundation}. For a stricter scenario where the retain set is not accessible during training, we opt for the \textbf{RWKU} benchmark, which focuses on famous individuals. In this case, we use Llama 3.1 8B instruct \cite{grattafiori2024llama3herdmodels}.

Finally, we evaluate the different methods on an \textbf{in-house e-commerce benchmark}, which reflects a real-world use case scenario.
The starting point is a model that has been trained on large amounts of public listing data from an e-commerce website.
The unlearning benchmark consists of three different unlearning targets, which are comprised by different sellers from the platform. 
Each seller has associated item listings, which need to be forgotten. %
The three sellers reflect three unlearning scenarios in the amount of data to be unlearned: a small, a medium and a large amount of listings.
Evaluation is performed by calculating ROUGE-recall score on text completions for forget and retain set respectively \cite{jin2024rwku}.\footnote{More details about the in-house benchmark will be available in the camera-ready version of the paper.}

More details on datasets and evaluation tasks used can be found in Appendix \ref{sec:ds}.

\textbf{Settings.} We closely follow the experimental setups described in the respective papers for each benchmark and cited method. All training is conducted using four A100 80GB GPUs.

For MUSE-News, we run each method for 10 epochs with $\lambda = 1$, a batch size of 32, and learning rates from the MUSE paper \cite{shi2024muse}. Additional hyperparameter tuning is performed to generate sweep curves, with reference ranges derived from the original method papers when available. Specifically, we use a learning rate of 5e-6 for ME+GD and 1e-5 for RKLD+KL.

For RWKU, we train for 3 epochs with varying learning rates (detailed in Section~\ref{sec:app_rwku}). Learning rate sweeps in the range [1e-7, 1e-5] are conducted for UnDIAL and Unilogit, while for other methods, we follow hyperparameter choices from their respective papers.

For our e-commerce benchmark, we train for 10 epochs for the smallest seller and 3 epochs for the medium and largest sellers. We begin with hyperparameters from the original method papers and continue tuning if necessary to achieve optimal results.

\subsection{Results on general unlearning scenario}
\begin{figure}[!t]
    \centering
    \includegraphics[width=0.48\textwidth]{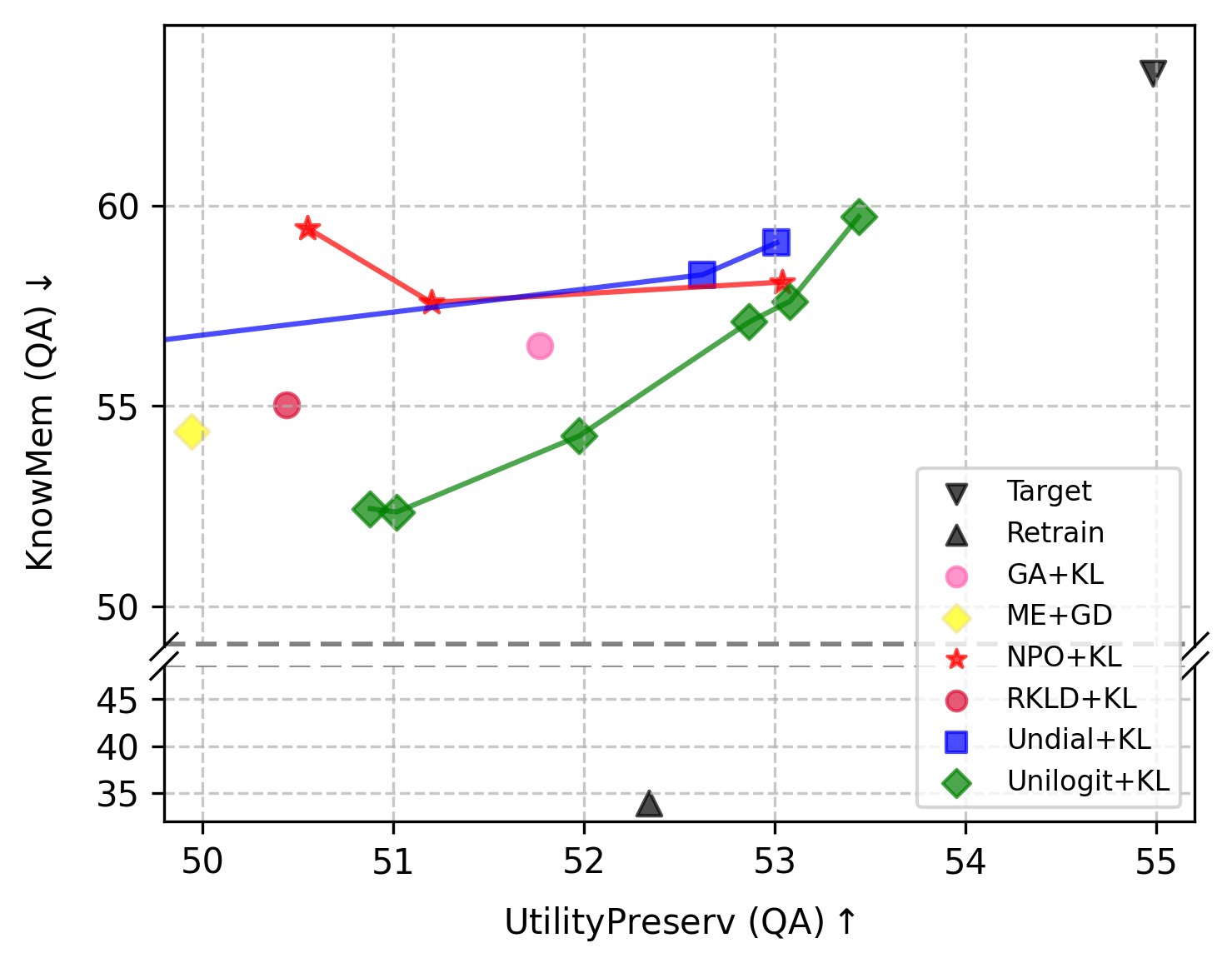}
    \caption{Results for the MUSE-News benchmark for different unlearning methods using multiple different hyperparameters. On the x-axis we have the retain performance and on the y-axis the forgetting performance, both for the QA task.}
    \label{fig:muse}
\end{figure}

In Figure \ref{fig:muse} we can see the results for the MUSE-News benchmark. %
We observe that out of the methods evaluated, Unilogit+KL, shows the most optimal Pareto curve, as opposed to the state-of-the-art NPO and its distillation-based competitor UnDIAL. This result substantiates our claim that Unilogit provides an effective and principled solution for unlearning, achieving superior trade-offs between forgetting and retention compared to existing methods.

Furthermore, the plot shows that our method is robust to hyperparameter tuning, providing a smooth, monotonically increasing curve when continuously varying the main unlearning hyperaprameter---the learning rate. In contrast, NPO and UnDIAL lack this smoothness, indicating greater sensitivity to hyperparameter choices. Specifically, for NPO, achieving optimal performance requires additional fine-tuning, such as adjusting the number of training epochs. This adds complexity to its deployment. Similarly, UnDIAL relies on multiple hyperparameters (learning rate and $\gamma$), making it challenging to tune effectively. Even with extensive tuning, UnDIAL ultimately results in suboptimal performance compared to Unilogit. The results for Undial visible on the plot are from learning rate 1e-5; however, Appendix \ref{sec:muse} contains a table with the full numbers for reference. The blue line, which continues off the plot is an interpolation of the UnDIAL results for learning rate 1e-4.

\begin{figure}[h]
    \centering
    \includegraphics[width=0.42\textwidth]{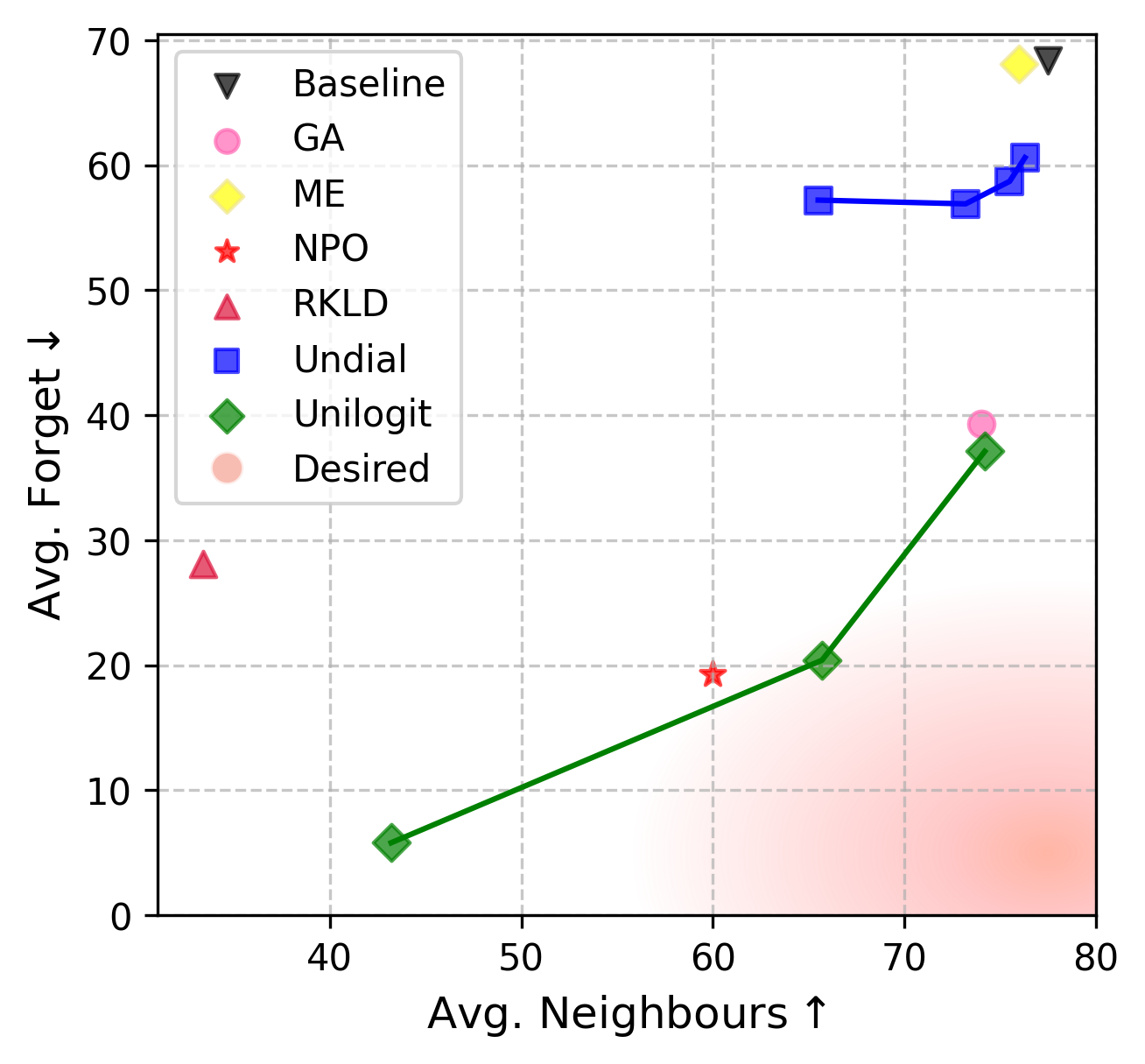}
    \caption{Results for the RWKU-News benchmark for different unlearning methods using multiple different hyperparameters. On the x-axis is the retain performance and on the y-axis the forgetting performance.}
    \label{fig:rwku}
\end{figure}
We further see that other baselines, such as GA+KL,  RKLD+KL and ME+DG result in a significant gap in unlearning efficacy for a similar level of utility preservation of Unilogit, underscoring the effectiveness of our method.

\subsection{Results on stricter unlearning scenario}
For the more difficult RWKU benchmark, where no retain set is available during unlearning, we present our result in Figure \ref{fig:rwku}. This setting lacks a golden retrain baseline for direct comparison. As a result, while we can still evaluate the Pareto-style curve optimality, we do not have a definitive reference point for maximum optimality (i.e., retrain model performance). To aid interpretation, we highlight in orange the plausible desired region of optimality, serving as a visual reference.

\begin{figure*}[!t]
    \centering
    \includegraphics[width=0.8\textwidth]{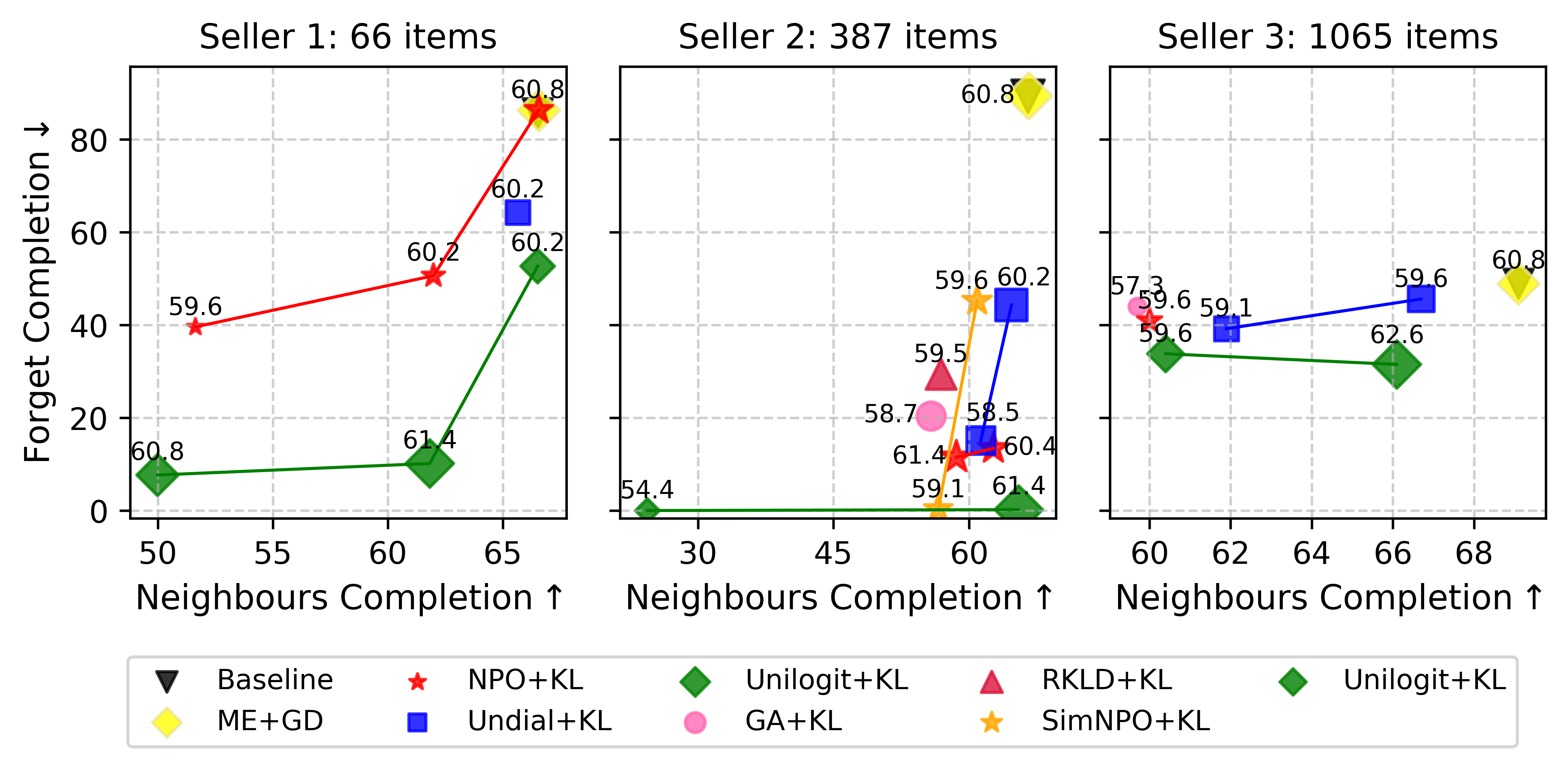}
    \caption{Comparison of unlearning methods on listings from three different sellers across three forget set sizes in our e-commerce dataset. Forget Completion and Neighbors Completion are evaluated using ROUGE-recall scores. Marker sizes and number annotations indicate MMLU scores, reflecting general model abilities.}
    \label{fig:internal}
\end{figure*}

In Figure \ref{fig:rwku}, Unilogit once again demonstrates superior performance over existing methods, as reflected in its monotonically-increasing curve, which is closest to the desired region. In contrast, UnDIAL exhibits inconsistencies, particularly at the $65.5$ Avg. Neighbours performance level, where it briefly shows higher Avg. Forget than at a nearby, higher-retain-performance setting. This suggests that UnDIAL reaches a local peak in hyperparameter space, particularly around the configuration (lr = 5e-6, $\gamma$ = 5), whereas Unilogit maintains Pareto optimality across its entire hyperparameter sweep.

NPO performs notably better on RWKU than UnDIAL, but still falls slightly short of Unilogit in terms of overall optimality. Interestingly, GA closely follows Unilogit, suggesting that in scenarios where GA has not yet undergone excessive forgetting (or catastrophic degradation), it could serve as a viable alternative. This highlights an important trend: performance differences between methods become most pronounced at higher levels of forgetting, reinforcing the need for robust, generalizable unlearning approaches.

\subsection{Results on in-house benchmark}
For our in-house e-commerce benchmark, we evaluate models using ROUGE-recall, the primary metric displayed in Figure \ref{fig:internal}. The figure plots ROUGE on the forgetting completion task against ROUGE on the completion task for the neighbor set of items. Across all three plots, Unilogit consistently achieves higher forgetting while maintaining similar or better retention performance compared to competing methods.

Each data point in Figure \ref{fig:internal} is annotated with an additional metric—general model utility, measured by MMLU accuracy \cite{hendryckstest2021mmlu} (higher is better). Notably, at comparable levels of retention, Unilogit+KL consistently achieves higher MMLU accuracy than other methods in all seller scenarios. The only exception is in the smallest-scale unlearning task (leftmost plot), where NPO+KL and ME+GD reach an MMLU accuracy of 60.8, slightly exceeding Unilogit's 60.2. However, this discrepancy is not indicative of superior performance, as both NPO+KL and ME+GD remain at baseline levels of forgetting, suggesting under-unlearning rather than a genuine trade-off advantage. This reinforces Unilogit's effectiveness in striking an optimal balance between unlearning and model utility preservation.

In the mid-scale seller scenario (middle plot in Figure \ref{fig:internal}), SimNPO+KL \cite{fan2024simplicity} matches Unilogit's Pareto frontier under specific conditions but lacks consistency. One configuration aligns with Unilogit's trade-off curve, while another exhibits lower retention despite only modest forgetting, a failure mode not observed in Unilogit. This instability highlights SimNPO+KL’s difficulty in maintaining an effective balance between forgetting and retention, whereas Unilogit+KL provides a good trade-off across hyperparameter settings.

These results underscore the practical reliability of Unilogit+KL, reaffirming its consistent ability to optimize forgetting while preserving model utility across diverse real-world scenarios. Its strong Pareto efficiency and hyperparameter robustness make it a suitable choice for unlearning tasks in production environments.

\begin{figure*}[htbp]
    \centering
    \begin{subfigure}[]{0.35\linewidth}
        \centering
        \includegraphics[width=\linewidth]{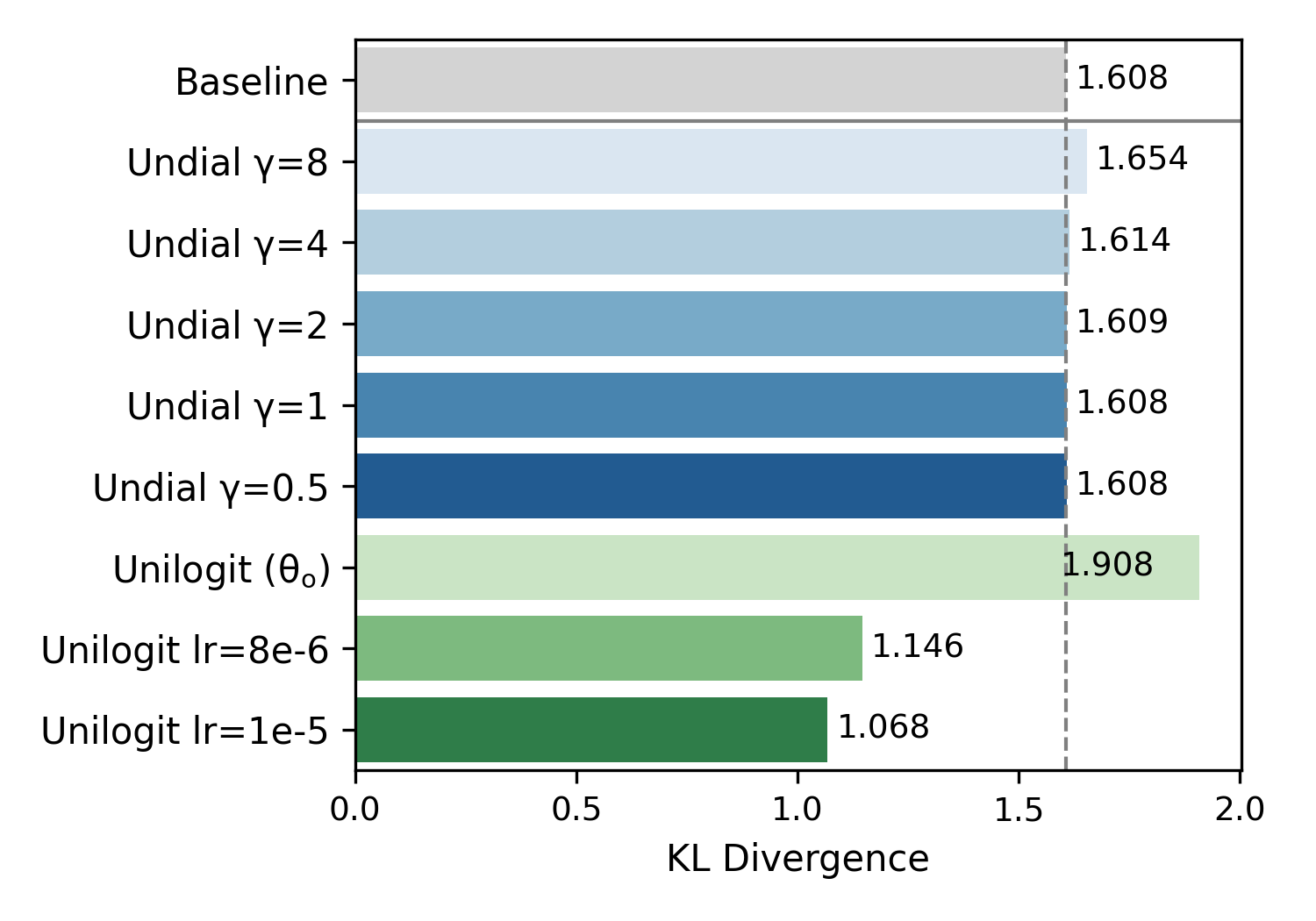}
        \label{fig:kl1}
    \end{subfigure}
    \hfill
    \begin{subfigure}[]{0.30\linewidth}
        \centering
        \includegraphics[width=\linewidth]{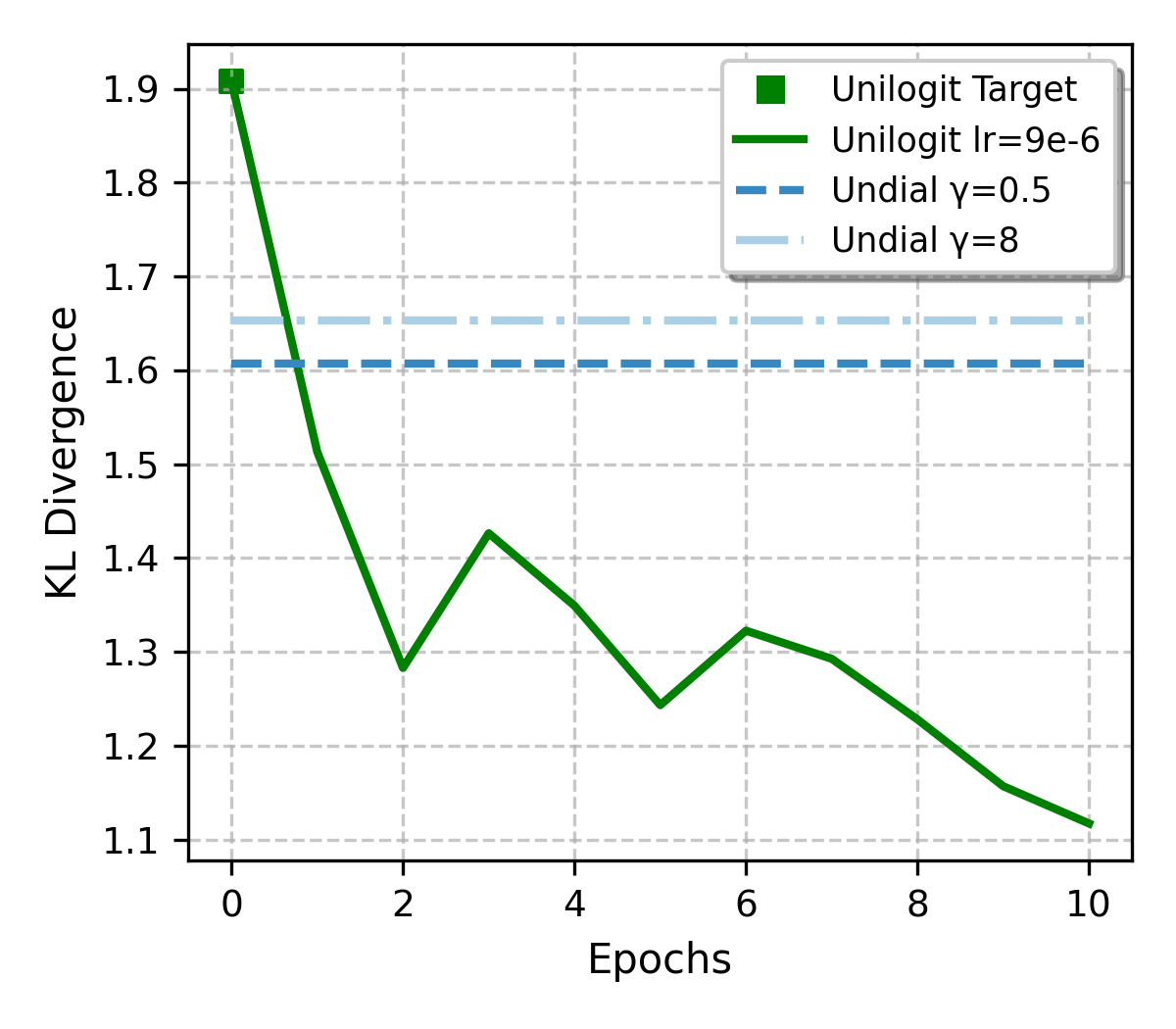}
        \label{fig:kl2}
    \end{subfigure}
    \hfill
    \begin{subfigure}[]{0.33\linewidth}
        \centering
        \includegraphics[width=\linewidth]{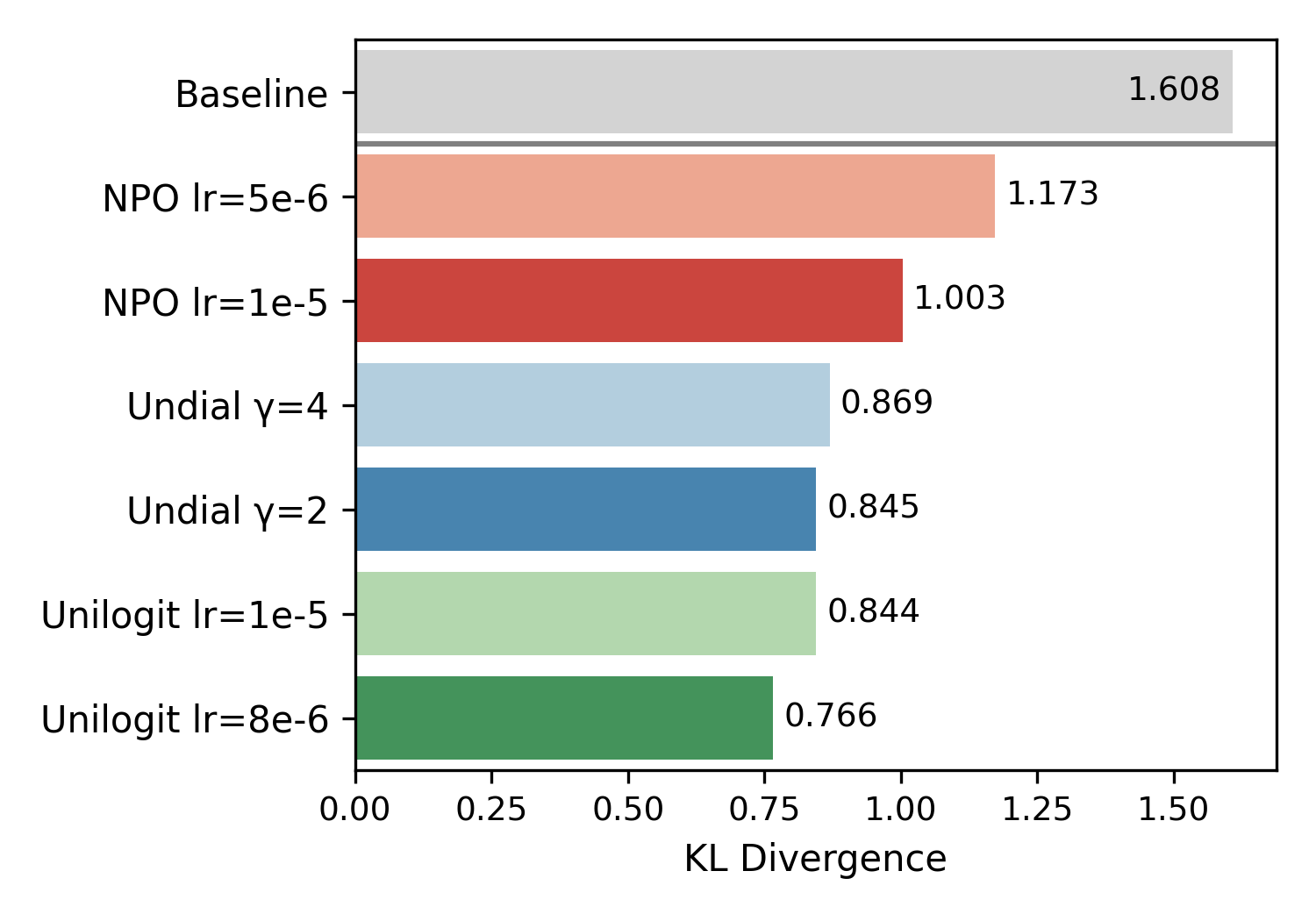}
        \label{fig:kl3}
    \end{subfigure}
    \caption{Left: Average KL divergence between the retrained model outputs and the \textit{soft labels} of UnDIAL and Unilogit on the forget set. Center: KL divergence progression between \textit{soft targets} and retrained model outputs for both methods over unlearning epochs. Right: Average KL divergence between unlearned model \textit{outputs} and retrained model for NPO, UnDIAL, and Unilogit. Lower values indicate better performance in all cases, as well as the baseline represents average KL between the \textit{outputs} of the starting model and the retrained model.}
    \label{fig:klevals}
\end{figure*}
\subsection{Unlearning target and output distribution analysis}
\label{sec:klevals} In this section we show that our method generates more accurate self-distillation targets than UnDIAL and produces output distributions after unlearning that more closely align with the retrain model than both NPO and UnDIAL on the forget set. To evaluate the accuracy of both the soft-label distributions and the final output distributions, we compute the KL divergence between these distributions and the retrain model’s outputs on 100 forget-set samples from MUSE-News.

In Figure \ref{fig:klevals} (left) is plotted the average KL divergence for the self-distillation targets $\tilde{p}(x;\theta)$ for UnDIAL and Unilogit between the respective unlearned models and the retrain model. The baseline on the plot is the average KL divergence between the outputs of the starting model ($\theta_o$) and the retrained model. We see that as we decrease the $\gamma$ parameter for Undial, we approach the baseline KL. That is intuitive because at $\gamma = 0$, the UnDIAL targets are just equal to the starting model outputs (Equation \ref{eq:undial}). We therefore find a drawback to the UnDIAL approach: if self-distillation targets rely solely on the original model, they cannot get closer in distance to the golden retrain model outputs. Achieving better alignment would possibly require a carefully tuned static $\gamma$ that works consistently across all samples. In contrast, Unilogit dynamically updates its targets using the current model state at each step. The last two bars in the plot, representing Unilogit’s final checkpoints with different learning rates, show significantly lower KL divergence than both UnDIAL and the baseline, demonstrating that Unilogit produces more accurate self-distillation targets.

Figure \ref{fig:klevals} (middle) further supports this, illustrating how Unilogit’s soft-label distributions become increasingly accurate throughout training, while UnDIAL’s by design remain static. As unlearning progresses, Unilogit’s targets exhibit decreasing KL divergence from the retrain model, reinforcing the advantage of dynamically updating targets based on the latest model state. If our assumption that a well-designed unlearning algorithm should bring the model closer to the retrain model at each step holds, then using the current model’s parameters to generate targets yields greater accuracy.

Finally, Figure \ref{fig:klevals} (right) assesses the KL divergence between the final output distributions of different unlearned models and the retrain model. This metric, as also seen in \citet{dong2024undialselfdistillationadjustedlogits}, captures the overall distributional alignment rather than focusing solely on individual predictions. The results indicate that Unilogit achieves a significantly closer match to the retrain model compared to NPO and UnDIAL, demonstrating that our method not only improves self-distillation targets but also better aligns the entire output distribution with the gold-standard retrain model.

\subsection{Ablations} \label{sec:ablations}
Our ablation experiments demonstrate that the use of Reverse KL divergence significantly improves forgetting performance while maintaining utility, and our method of calculating target logit values outperforms UnDIAL. Additionally, using the latest model weights results in a more optimal model. These findings confirm that the key features of our approach---soft label calculation, reverse KL loss, and current model logits---are crucial for enhancing unlearning performance. For a detailed discussion, please refer to Appendix \ref{sec:ablations}.

\section{Conclusion}
We present Unilogit, a novel method for efficient and effective machine unlearning. Through extensive experimentation across multiple benchmarks, we demonstrate that Unilogit outperforms existing state-of-the-art methods, 
\begin{figure}[!h]
    \centering
    \includegraphics[width=0.45\textwidth]{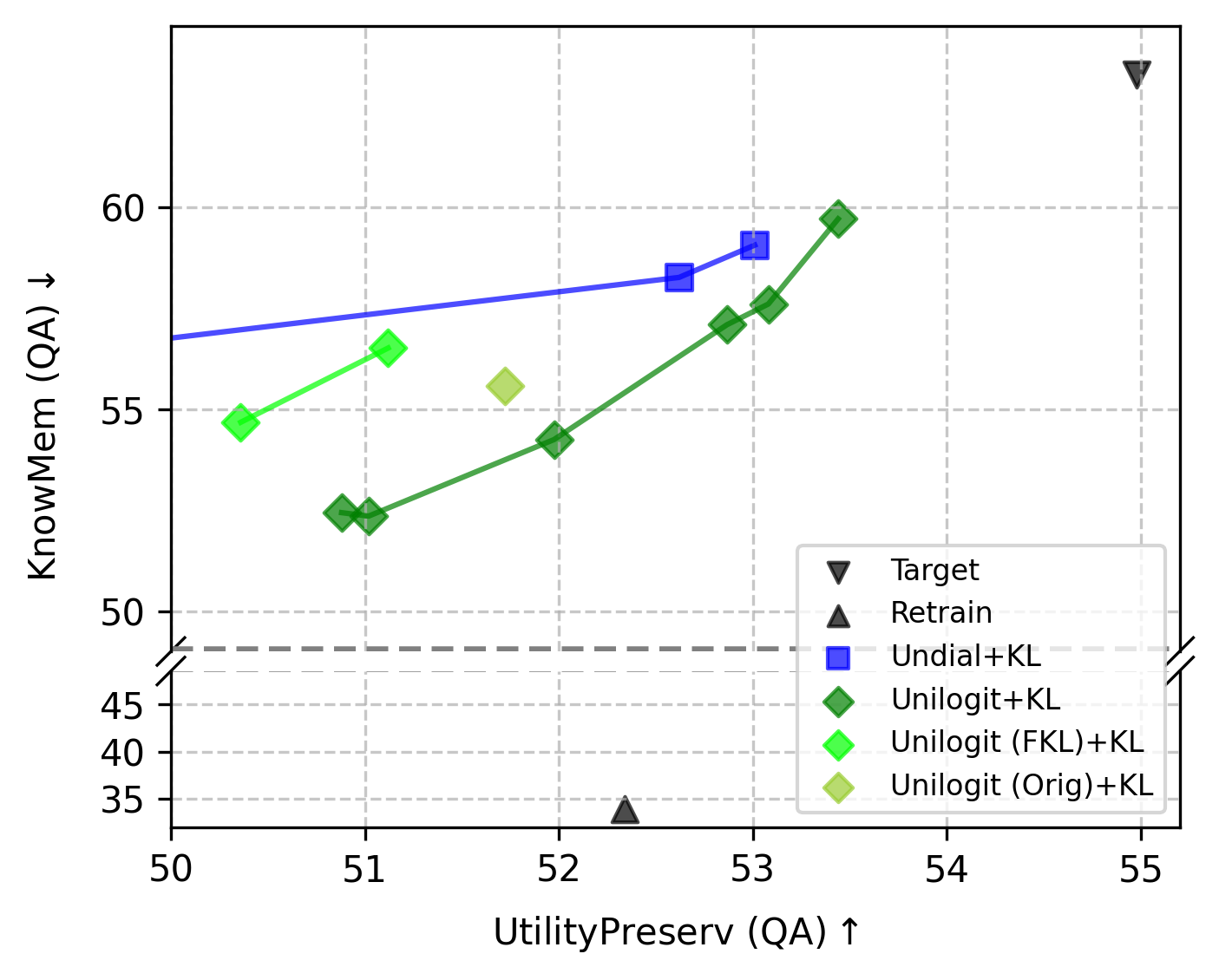}
    \caption{Results of Unilogit ablations on MUSE-News. Unilogit with 1) FKL loss and 2) original model distillation targets were tuned to match Unilogit+KL in UtilityPreserv, allowing comparison on KnowMem performance. Unidial+KL is included for reference.}
    \label{fig:ablations}
\end{figure}
in terms of both forgetting
and utility preservation. Our results show that Unilogit achieves the most optimal Pareto curves, with superior performance in retaining model utility while effectively forgetting unwanted knowledge. Furthermore, Unilogit exhibits robustness to hyperparameter tuning, providing stable and consistent performance across various settings.

We also provide an in-depth analysis of Unilogit’s self-distillation targets and output distributions, highlighting its ability to generate more accurate targets compared to UnDIAL and its capacity to maintain better output distributions post-unlearning. The ablation studies reinforced the importance of key design choices, such as using Reverse KL divergence, calculating soft labels dynamically, and leveraging current model logits during training, all of which contribute to Unilogit’s superior performance.

Overall, Unilogit demonstrates both practical 
robustness and theoretical advantages, making it 
a promising method for real-world applications.

\section*{Limitations}
Although Unilogit has been tested on multiple datasets, further evaluation across an even broader range of benchmarks is essential to fully understand its capabilities and limitations. While we have purposefully selected a diverse set of evaluation benchmarks to capture a variety of scenarios, more evaluations for additional domains should be performed.

Secondly, Unilogit currently does not account for the varying importance or relevance of each token in the context of the information to be forgotten. The method treats all tokens equally during the unlearning process, which might not be optimal for situations where certain tokens play a more critical role in the forgetting objective, such as production settings with structured text. Introducing a mechanism to weigh tokens based on their significance could enhance the precision of the unlearning process. Addressing this issue would require introducing the appropriate intrinsic bias, presenting an intriguing avenue for future research. This development could serve as an additional tool for more effective targeted unlearning, achieving a better balance between forgetting and retaining information.

Additionally, in the current study we focus only on the English language and on models with a size around 7-8 billion parameters.
Future work should take more languages and different model sizes into account.

By acknowledging these limitations, we aim to highlight areas where Unilogit can be refined and improved, paving the way for future research.

\section*{Ethical Considerations}
In the development and deployment of Unilogit, we acknowledge the ethical imperative to prioritize user privacy and data protection. Our method aims to address the critical need for machine unlearning, ensuring compliance with regulations such as the GDPR's "right to be forgotten." By effectively removing sensitive information from language models, Unilogit helps mitigate privacy risks associated with data leakage and unauthorized data retention. However, it is essential to consider potential misuse, such as the erasure of accountability in decision-making systems. Therefore, we advocate for responsible implementation, ensuring that unlearning is applied judiciously and in contexts where it enhances user rights and data security without compromising ethical standards.

\bibliography{acl_latex}

\appendix

\section{Extended Related Work} \label{sec:app_related}
Among parameter-tuning methods, \textbf{Gradient Ascent (GA)} \cite{jang-etal-2023-knowledge} serves as a fundamental baseline, applying gradient ascent on the forget set to reverse learned representations. However, GA is highly unstable, leading to catastrophic forgetting of retain knowledge \cite{zhang2024negative}.

To address this, \textbf{Negative Preference Optimization (NPO)} \cite{zhang2024negative} introduces a loss function derived from preference optimization theory \cite{rafailov2024direct}, using only the forget samples as negative samples. It controls the forgetting process using a hyperparameter, which prevents the model from diverging into instability. As a result, NPO has been recognized as a robust state-of-the-art method and baseline for future research \cite{wang2024rkldreversekldivergencebasedknowledge, fan2024simplicity, dong2024undialselfdistillationadjustedlogits}. The objective function for NPO is:
\begin{multline*}
    \label{eq:npo}
    \mathcal{L}_{\mathrm{NPO}, \beta}(\theta) = \\
    -\frac{2}{\beta} \mathbb{E}_{D_f}\left[\log \sigma\left(-\beta \log \frac{p(y \mid x; \theta)}{p(y \mid x; \theta_{o})}\right)\right]
\end{multline*}

Another approach, \textbf{ME+GD} \cite{uniformce}, maximizes the entropy of the model’s predictions on the forget set by pushing output probabilities towards a uniform distribution. This method directly relates to our main research question (\ref{rq:question}), as it establishes a principled lower-bound baseline---using the uniform target as the least assumptive distribution for unlearning knowledge. Unlike the commonly used KL-divergence regularizer, ME+GD employs cross-entropy with one-hot labels to preserve performance on the retain set. The objective for ME+GD is:
\begin{align*}
    &\mathcal{L}_{ME+GD}(\theta, D_f, D_r) \\
    =&\mathbb{E}_{(x, y) \in D_f} \text{KL}(p(y|x;\theta) \, || \, \mathcal{U}_{|V|}) \, + \\
    &\lambda \, \mathbb{E}_{(x,y) \in D_r} H(p(y|x),y) \\
    \equiv &-\mathbb{E}_{(x,y) \in D_r}H(p(y|x;\theta)) \,+ \\
    &\lambda \, \mathbb{E}_{(x,y) \in D_r} H(p(y|x),y)
\end{align*}

\citet{eldan2023whosharrypotterapproximate} were among the first to introduce the idea of using an \textit{auxiliary model} to facilitate the unlearning process. Their method, \textbf{"Who is Harry Potter" (WHP)}, involves fine-tuning the original model on the forget set to create a \textit{"reinforced" model}. This reinforced model is then used to extract distributional information about the tokens most associated with the knowledge being unlearned. Using this information, the authors propose an output mixing equation to generate fine-tuning labels for unlearning.

The concept of leveraging an auxiliary model aligns naturally with \textit{knowledge distillation} (KD) \cite{hintonkd}, making WHP a foundational approach that has inspired subsequent \textbf{distillation-based unlearning} methods \cite{wang2024rkldreversekldivergencebasedknowledge, dong2024undialselfdistillationadjustedlogits, chen-yang-2023-unlearnacl, chundawat2023can}. One such method, \textbf{RKLD} \cite{wang2024rkldreversekldivergencebasedknowledge} refines the WHP approach by enhancing model reinforcement and distribution mixing, then distilling the modified soft labels into the original model to achieve targeted unlearning while preserving overall performance. To calculate the target logits for later distilling them onto the original model in RKLD \cite{wang2024rkldreversekldivergencebasedknowledge}, the following equation is used:
\begin{align*}
&h_\text{RKLD}(x;\theta)= \\
&h(x;\theta_o)-\alpha \text{ReLU}(h(x;\theta_s)-h(x;\theta_o))    
\end{align*}
Here, $\theta_s$ are the parameters of the strengthened (reinforced) model in the method.

Building on the idea of leveraging modified target distributions for unlearning, a more recent method, \textbf{UnDIAL} \cite{dong2024undialselfdistillationadjustedlogits}, introduces a self-distillation approach that adjusts model logits to achieve forgetting. Inspired by this direction, our proposal refines the process of crafting effective soft labels for unlearning. UnDIAL generates its self-distillation targets by applying softmax to adjusted logits, where the original model logits $h(x; \theta)$ are modified by reducing the target logit's value, controlled by the hyperparameter $\gamma$:
\begin{equation} \label{eq:undial}
    \tilde{h}_\text{UnDIAL}(x;\theta)= h(y|x;\theta)-\gamma \mathbf{t} \,,
\end{equation}
where $\mathbf{t}$ is the one-hot encoded target vector.

\section{Deriving the target distribution for Unilogit}
\label{appendix:derivation_unilogit}
As outlined in Section \ref{sec:methods}, our goal with calculating the target logits is to have a uniform distribution value of $1/|V|$ of the target logit after the softmax. To achieve this, we can derive the exact logit value that would result in the desired outcome. With $\bm{\tilde{h}}_k$ we denote the unknown value of the target logit with index $k$. Starting with what we know:

\begin{align*}
    \frac{e^{\bm{\tilde{h}}_k(x;\theta)}}{e^{\bm{\tilde{h}}_k(x;\theta)} + \sum\limits_{i \neq k}^{|V|} e^{\bm{h}_i(x;\theta)}} = \frac{1}{|V|} \\
    e^{\bm{\tilde{h}}_k(x;\theta)} = \frac{1}{|V|} \left(e^{\bm{\tilde{h}}_k(x;\theta)} + \sum\limits_{i \neq k}^{|V|} e^{\bm{h}_i(x;\theta)} \right) \\
    e^{\bm{\tilde{h}}_k(x;\theta)}\left( 1-\frac{1}{|V|} \right) = \frac{\sum\limits_{i \neq k}^{|V|} e^{\bm{h}_i(x;\theta)}}{|V|} \\
    e^{\bm{\tilde{h}}_k(x;\theta)} = \frac{\sum\limits_{i \neq k}^{|V|} e^{\bm{h}_i(x;\theta)}}{|V|-1} \\
\end{align*}
Taking the log of both sides:
\begin{align*}
    \bm{\tilde{h}}_k(x;\theta)= \log \frac{\sum\limits_{i \neq k}^{|V|} e^{\bm{h}_i(x;\theta)}}{|V|-1}
\end{align*}
This gives us the required value of the logit so that after the softmax we get a uniform probability for target token $k$. We can then vectorize the equation for calculating the target logits:
\begin{align*}
    \bm{\tilde{h}}(x; \theta) = \, &
    (1-\bm{t})\, \bm{h}(x; \theta) \,+ \\ 
    & \bm{t} \, \log \sum\limits_{i \neq k}^{|V|}\frac{\exp({\bm{h}_i{(x;\theta})}) }{|V|-1}   
\end{align*}
where $\bm{t}$ is the target one-hot labels for a given input sample $x$ from the forget set. This way, we will only update the value of the $k$-th logit in the modified logits vector $\bm{\tilde{h}}(x;\theta)$.

\section{Experiments Details}
\subsection{Datasets and Models} \label{sec:ds}
\subsubsection{MUSE-News} \label{sec:muse}
The \textbf{MUSE-News} benchmark \cite{shi2024muse} as our general unlearning scenario benchmark. It consists of BBC News articles. The evaluation tasks include question-answering type sets for forget knowledge (KnowMem.) and retain knowledge (UtilityPreserv.), a membership inference test (PrivLeak) and a text completion task for the forget data (VerbMem).

The MUSE-News dataset consists of BBC News articles published after August 2023, ensuring that the pre-trained Llama 2 7B model has not encountered any of the articles. The dataset is split into a Forget Set (3554 passages) and a Retain Set (3555 passages). The target model is trained on both sets, while the retrained model is trained only on the retain subset. Unlearning is performed on a subset of 889 articles from the forget set, and the model is evaluated on tasks such as Verbatim Memorization, Knowledge Memorization (QA), Privacy Leakage, and Utility Preservation (QA), as outlined by the MUSE-News benchmark \cite{shi2024muse}. 

VerbMem is a text completion task, where ROUGE-F1 is used to measure completion performance. Knowledge Memorization and Utility Preservation are both question-answering type tasks, where ROUGE is also employed to assess answer accuracy. Finally, Privacy Leakage is a membership inference test, as measured by Min-K\% Prob \cite{shi2023mink}.

\subsubsection{RWKU Benchmark}
As a stricter unlearning scenario without access to retain set examples during unlearning, we adopt the \textbf{RWKU} benchmark \cite{jin2024rwku}. It is a compilation of knowledge about each of the 100 most famous people, according to Wikipedia. As starting model we use Llama 3.1 8B Instruct \cite{grattafiori2024llama3herdmodels}, whose pre-training already includes information about all the people in the dataset, so we can start unlearning from that checkpoint. The evaluation tasks for the forget set include fill-in-the-blank style samples, question-answering and adversarial attack samples, ordered by difficulty. For the retain set, there are fill-in-the blank and QA-style samples. Furthermore, the benchmark includes a MIA attack evaluation and general utility evaluations: MMLU \cite{hendryckstest2021mmlu}, BIG-Bench Hard \cite{suzgun2022challenging}, TruthfulQA \cite{lin-etal-2022-truthfulqa}, TriviaQA \cite{joshi-etal-2017-triviaqa}, Alpaca Eval (Fluency) \cite{alpacaeval}.

\subsubsection{Internal e-commerce benchmark}
In our e-commerce experiments, we evaluate several tasks to comprehensively assess model performance: the completion task, prediction probability (loss), and general utility, which includes MMLU and an e-commerce-specific task.

The completion task involves measuring the ROUGE-recall score for completing the second half of a seller's item description, given the first half. This setup, similar to the VerbMem task in MUSE \cite{shi2024muse}, is designed to evaluate the model's ability to recall text accurately on a word-by-word basis. We calculate this metric for both the retain and forget sets.

The prediction probability metric assesses the log-likelihood of a given sample item, providing insight into the model's confidence in its predictions.

The internal e-commerce task evaluates the model's ability to answer fill-in-the-blank style queries about an item, testing its understanding and retention of specific item details.

Together, these tasks offer a comprehensive evaluation of the model's forgetting and retention performance, as well as its overall utility in an e-commerce context.

\section{Extra Results}

\subsection{Ablations} \label{sec:ablations}
In our ablation experiments we evaluate core design aspects of our method. First, we ablate the choice of using Reverse KL divergence as forgetting optimization objective. As explained in Section \ref{sec:unilogit}, RKL is more suitable for the unlearning problem in this setting. Despite that, we add a quantitative argument to support our choice. In Figure \ref{fig:ablations} we can clearly see that for the same level of utility preservation, we get significantly better forgetting performance for RKL. 

We carry out a second ablation where, likewise to UnDIAL, we create our self-distillation targets using the starting model. In that sense, Unilogit (Orig)+KL and UnDIAL differ only by their respective processes of logit diminishment. On Figure \ref{fig:ablations} we see that this version of Unilogit outperforms UnDIAL, which demonstrates that our methodology of calculating the target logit value is more effective than UnDIAL.

If we make the comparison between Standard Unilogit+KL and Unilogit (Orig)+KL we see that using the latest model weights yields a more optimal model. 

Ultimately, the ablation results show that the three most important features of our method---the soft label calculation, the reverse KL loss and using the current model logits are all beneficial to unlearning performance.  

\subsection{MUSE Benchmark}
In Table \ref{tab:muse}, we see the full results from all our runs on MUSE-News.

\begin{table*}[ht]
    \centering
    \begin{tabular}{llcccccc}
        \toprule
        \textbf{Method} & \textbf{Hyperparameters} & \textbf{VerbMem} & \textbf{KnowMem} & \textbf{PrivLeak} & \textbf{UtilityPreserv} \\
        & & \textbf{(completion) $\downarrow$} & \textbf{(QA) $\downarrow$} & & \textbf{(QA) $\uparrow$} \\
        \midrule
        Target & & 56.64 & 63.30  & -99.81 & 54.98 \\
        Retain & & 20.30 & 33.93  & 0.00 & 52.34 \\
        \midrule
        NPO & lr=1e-5 & 7.24 & 7.24 & 57.40 & 7.74 & \\
        NPO+KL & lr=1e-5, $\beta$=0.1 & 40.71 & 59.44 & -93.76 & 50.55 \\
        NPO+KL & lr=5e-6, $\beta$=0.1 & 47.70 & 58.09 & -98.09 & 53.04 \\
        NPO+KL & lr=5e-6, $\beta$=0.1 & 47.01 & 57.58 & -98.09 & 51.20 \\
        GA+KL & lr=1e-5 & 41.22 & 56.51 & -99.75 & 51.77 \\
        RKLD+KL & lr=1e-5 & 47.96 & 55.02 & -99.75 & 50.44 \\
        ME+GD & lr=5e-6  & 40.23 & 	54.37  & 	-99.75 &	49.95 \\
        Undial+KL & lr=1e-5, $\gamma$=2 & 42.37 & 58.27 & -99.75 & 52.62 \\
        Undial+KL & lr=1e-5, $\gamma$=4 & 41.52 & 59.07 & -99.75 & 53.01 \\
        Undial+KL &	lr=1e-5, $\gamma$=8	& 40.81 &	55.97	& -99.75	& 48.60 \\
        Undial+KL & lr=1e-4, $\gamma$=4 & 21.15 & 18.69 & -96.94 & 26.47 \\
        Undial+KL & lr=1e-4 $\gamma$=2	& 22.63 &	25.88	&	-98.11	& 30.78 \\
        \midrule
        Unilogit+KL & lr=1e-5 & 39.35 & 52.44 & -99.75 & 50.88 \\
        Unilogit+KL & lr=1e-5 & 33.60 & 52.46 & -99.71 & 48.15 \\
        Unilogit+KL & lr=5e-6 & 51.61 & 59.72 & -99.79 & 53.44 \\
        Unilogit+KL & lr=8e-6 & 43.01 & 57.61 & -99.77 & 53.08 \\
        Unilogit+KL &	lr 8.5e-6 &	41.38	 & 57.10 &	-99.77 & 52.87 \\
        Unilogit+KL &	lr 8.75e-6 &	39.54 &	54.25	&	-99.75 &	51.98 \\
        \bottomrule
    \end{tabular}
    \caption{Results of various methods on MUSE-News on Llama 2 7B}
    \label{tab:muse}
\end{table*}

\subsection{RWKU Benchmark} \label{sec:app_rwku}
In Table \ref{tab:rwku}, we see the full results from all our runs on RWKU. 

\begin{table*}[]
\small
\centering
\setlength{\tabcolsep}{5.5pt}
\begin{tabular}{@{}c|c|cccc|ccc|cc|ccccc@{}}
\toprule
\multirow{2}{*}[-2pt]{\textbf{Method}} & \multirow{2}{*}[-2pt]{LR} & \multicolumn{4}{c|}{\textbf{Forget Set}} & \multicolumn{3}{c|}{\textbf{Neighbor Set}} & \multicolumn{2}{c|}{\textbf{MIA Set}} & \multicolumn{5}{c}{\textbf{General Benchmarks}} \\ \cmidrule(l){3-16} 

                                                    &                                                                              & FB       & QA       & AA       & All     & FB           & QA           & All          & FM                & RM                & Gen      & Rea     & Tru     & Fac     & Flu    \\ \midrule
Baseline                                                                                                   & -                                                                            & 65.0     & 69.9     & 70.3     & 68.4    & 76.4         & 78.5         & 77.5         & 2.1               & 2.2               & 66.5     & 43.2    & 36.4    & 62.3    & 7.0    \\ \midrule
GA                                                                                                         &     2e-6                                                                         & 37.5     & 33.8     & 46.6     & 39.3    & 73.6         & 74.4         & 74.0         & 2.4               & 2.2               & 66.2     & 43.0    & 36.7    & 62.8    & 6.9    \\
RKLD                                                                                                       &     1e-5                                                                         & 25.6     & 28.0     & 30.9     & 28.1    & 36.6         & 30.1         & 33.4         & 13.8              & 11.3              & 63.1     & 20.5    & 35.6    & 13.9    & 6.3    \\ 
ME	& 3e-6 &	64.4 &	70.0 &	69.8 &	68.1 &	74.5 &	77.5 &	76.0 &	2.1 &	2.2 &	66.2 &	43.2 &	35.5 &	62.4 &	7.0 \\
NPO                                                                                                        &       2e-6                                                                       & 24.0     & 16.1     & 17.9     & 19.3    & 57.9         & 62.0         & 60.0         & 2.6               & 2.3               & 65.9     & 43.2    & 35.3    & 63.2    & 6.7    \\
Undial                                                                                                     & 2e-7                                                         & 57.7     & 59.9     & 64.1     & 60.6    & 74.4         & 78.3         & 76.3         & 2.2               & 2.2               & 66.1     & 43.4    & 35.6    & 62.9    & 7.0    \\
Undial                                                                                                     & 6e-7                                                         & 57.2     & 56.5     & 62.4     & 58.7    & 73.6         & 77.4         & 75.5         & 2.2               & 2.2               & 66.1     & 43.4    & 35.3    & 62.7    & 6.9    \\
Undial                                                                                                     & 5e-6                                                          & 56.3     & 53.5     & 60.8     & 56.9    & 71.9         & 74.5         & 73.2         & 2.2               & 2.2               & 66.0     & 41.9    & 36.1    & 59.8    & 6.9    \\
Undial                                                                                                     & 1e-5                                                        & 56.9     & 54.4     & 60.3     & 57.2    & 62.1         & 69.0         & 65.5         & 2.3               & 2.3               & 65.0     & 41.0    & 35.4    & 48.5    & 6.6    \\ \midrule
Unilogit                                                                                                   & 5e-7                                                                      & 5.7      & 3.7      & 8.1      & 5.8     & 44.8         & 41.7         & 43.2         & 7.0               & 4.4               & 65.2     & 28.1    & 29.1    & 57.0    & 5.8    \\
Unilogit                                                                                                   & 2e-7                                                                      & 20.5     & 15.4     & 25.4     & 20.4    & 63.8         & 67.5         & 65.7         & 3.4               & 2.4               & 65.7     & 41.9    & 35.1    & 63.3    & 6.7    \\ \midrule
\end{tabular}
\caption{Results of various methods for unlearning on the RWKU benchmark on Llama 3.1 8B.}
\label{tab:rwku}
\end{table*}

\subsection{In-House Dataset Unlearning Task}
In Tables \ref{tab:ebay1}-\ref{tab:ebay3} we show the full tables for the results on our in-house e-commerce benchmark for unlearning. It consists of three entities with associated structured passages to them. We run unlearning on each of these entities for each unlearning method.

\begin{table*}[]
\centering
\begin{tabular}{l|cc|cc|cc}
\toprule
\multirow{2}{*}{Method} & \multicolumn{2}{c|}{Forget Set} & \multicolumn{2}{c|}{Neighbours} & \multirow{2}{*}{MMLU $ \uparrow$} & \multirow{2}{*}{e-Commerce Task $\uparrow$} \\  
                        & Rouge $\downarrow$ & Loss $\uparrow$  & Rouge $\uparrow$ & Loss $\downarrow$  &                            &                               \\ \midrule
Baseline & 85.6                     & 0.72                     & 72.0                     & 0.52                      & 60.8                                            & 55.2                            \\ \midrule
NPO+KL                           & 39.6                     & 1.81                     & 58.8                     & 0.75                      & 59.7                                            & 53.9                            
            \\ 
NPO+KL                           & 50.6                     & 1.86                     & 67.1                     & 0.74                      & 60.2                                            & 53.9                            \\ 
ME+GD                            & 86.4                     & 0.72                     & 72.5                     & 0.52                      & 60.8                                            & 55.2                            \\
Undial+KL                        & 64.3                     & 0.90                     & 69.0                     & 0.55                      & 60.3                                            & 54.9                            \\
Unilogit+KL                      & 7.7                      & 6.16                     & 51.6                     & 0.90                      & 60.8                                            & 53.6                            \\ \midrule
Unilogit+KL                      & 10.2                     & 6.84                     & 62.5                     & 0.82                      & 61.4                                            & 54.3                            \\
Unilogit+KL                      & 52.7                     & 1.23                     & 71.6                     & 0.54                      & 60.3                                            & 55.1   \\ \bottomrule                        
\end{tabular}
\caption{Results of internal e-commerce benchmark for seller with 66 items (small-scale seller) on Llama 3.1 8B. }
\label{tab:ebay1}
\end{table*}

\begin{table*}[]
\centering
\begin{tabular}{l|cc|cc|cc}
\toprule
\multirow{2}{*}{Method} & \multicolumn{2}{c|}{Forget Set} & \multicolumn{2}{c|}{Neighbours} & \multirow{2}{*}{MMLU $ \uparrow$} & \multirow{2}{*}{e-Commerce Task $\uparrow$} \\  
                        & Rouge $\downarrow$ & Loss $\uparrow$  & Rouge $\uparrow$ & Loss $\downarrow$  &                            &                               \\ \midrule
Baseline    & 89.3                             & 0.10                            & 80.8                                 & 0.28                                & 60.8                     & 54.3                             \\ \midrule
NPO         & 13.1                             & 1.06                            & 57.9                                 & 0.52                                & 60.2                     & 51.8                             \\
GA          & 11.6                             & 5.03                            & 66.7                                 & 0.33                                & 59.6                     & 53.1                             \\ \midrule
GA+KL       & 20.4                             & 0.47                            & 67.5                                 & 0.33                                & 58.7                     & 53.4                             \\
NPO+KL      & 11.5                             & 0.89                            & 69.0                                   & 0.33                                & 61.4                     & 53.0                               \\
NPO+KL      & 13.4                             & 0.78                            & 75.0                                   & 0.31                                & 60.4                     & 53.7                             \\
RKLD+KL     & 29.4                             & 0.17                             & 66.1                                 & 0.34                                & 59.5                     & 53.8                             \\
SimNPO+KL   & 0.3                              & 33.63                           & 68.1                                 & 0.35                                & 59.1                     & 52.6                             \\
SimNPO+KL   & 45.2                             & 0.26                             & 72.7                                 & 0.31                                & 59.6                     & 53.5                             \\
ME+GD       & 89.4                             & 0.10                            & 80.7                                 & 0.28                                & 60.8                     & 54.3                             \\ 
UnDIAL+KL      & 44.3                             & 0.17                            & 79.0                                   & 0.28                                & 60.2                     & 54.2                             \\
UnDIAL+KL      & 15.1                             & 0.27                            & 75.9                                 & 0.29                                & 58.5                     & 54.1                             \\
\midrule
Unilogit+KL & 0.0                                & 10.78                           & 25.8                                 & 1.68                                & 54.4                     & 43.4                             \\
Unilogit+KL & 0.2                              & 6.64                            & 78.2                                 & 0.29                                & 61.4                     & 53.8                             \\
\bottomrule
\end{tabular}
\caption{Results of internal e-commerce benchmark for seller with 387 items (medium-scale seller) on Llama 3.1 8B. }
\label{tab:ebay2}
\end{table*}

\begin{table*}[]
\centering
\begin{tabular}{l|cc|cc|cc}
\toprule
\multirow{2}{*}{Method} & \multicolumn{2}{c|}{Forget Set} & \multicolumn{2}{c|}{Neighbours} & \multirow{2}{*}{MMLU $ \uparrow$} & \multirow{2}{*}{e-Commerce Task $\uparrow$} \\  
                        & Rouge $\downarrow$ & Loss $\uparrow$  & Rouge $\uparrow$ & Loss $\downarrow$  &                            &                               \\ \midrule
Baseline    & 48.9                             & 0.54                            & 58.1                                 & 0.54                                & 60.8                     & 53.8                             \\ \midrule
GA+KL       & 44.0                             & 0.41                            & 52.5                                 & 0.62                                & 57.3                     & 53.3                             \\
NPO+KL      & 41.0                             & 1.00                            & 51.7                                 & 0.70                                & 59.6                     &  53.8            \\
ME+GD       & 48.9                             & 0.54                            & 58.2                                 & 0.54                                & 60.8                     & 53.8                             \\
Undial+KL   & 45.6                             & 0.66                            & 55.8                                 & 0.59                                & 59.6                     & 53.8                             \\
Undial+KL   & 39.2                             & 0.78                            & 50.2                                 & 0.64                                & 59.1                     & 53.5                             \\  \midrule
Unilogit+KL & 33.8                             & 1.06                            & 46.8                                 & 0.69                                & 59.6                     & 53.4                             \\
Unilogit+KL & 31.5                             & 3.86                            & 55.6                                 & 0.60                                & 62.6                     & 53.9  \\
\bottomrule    
\end{tabular}
\caption{Results of internal e-commerce benchmark for seller with 1065 items (large-scale seller) on Llama 3.1 8B. }
\label{tab:ebay3}
\end{table*}

\end{document}